%% file: naaclhlt2019.tex
\documentclass[11pt,a4paper]{article}
\usepackage{authblk}
\usepackage[hyperref]{naaclhlt2019}
\usepackage{times}
\usepackage{latexsym}
\usepackage{amsfonts}
\usepackage{amsmath}

\usepackage{url}

\aclfinalcopy 
\def\aclpaperid{1748} 

\usepackage{hyperref}
\usepackage{url}

\usepackage{pgfplots}
\usepackage[caption=false]{subfig}
\usepackage{breqn}

\DeclareMathOperator{\argmax}{argmax}

\newcommand{\slmodel}{{\textsc {Shortlister}}}
\newcommand{\newmodel}{{\textsc {CoNDA}}}

\newcommand{\norm}[1]{\left\lVert#1\right\rVert}

\newtheorem{definition}{Definition}
\newtheorem{example}{Example}

\definecolor{belize_hole_blue}{rgb}{0.1607843137254902,0.5019607843137255,0.7254901960784313}
\definecolor{alizarin_red}{rgb}{0.90588, 0.29803921, 0.235294117647059}
\definecolor{nephritis_green}{rgb}{0.15294117647058825,0.6823529411764706,0.3764705882352941}
\definecolor{orange}{rgb}{0.9529411764705882,0.611764705882353,0.07058823529411765}
\definecolor{wisteria_purple}{rgb}{0.5568627450980392,0.26666666666666666,0.6784313725490196}
\definecolor{concrete_gray}{rgb}{0.5843137254901961,0.6470588235294118,0.6509803921568628}

\title{Continuous Learning for Large-scale Personalized Domain Classification}

\author[1]{\textbf{Han Li}}
\author[2]{\textbf{Jihwan Lee}}
\author[2]{\textbf{Sidharth Mudgal}}
\author[2]{\textbf{Ruhi Sarikaya}}
\author[2]{\textbf{Young-Bum Kim}}
\affil[1]{University of Wisconsin, Madison}
\affil[ ]{\texttt{hanli@cs.wisc.edu}}
\affil[2]{Amazon Alexa AI}
\affil[ ]{\texttt{\{jihwl, sidmsk, rsarikay, youngbum\}@amazon.edu}}

\date{}

\begin{document}
\maketitle
\begin{abstract}
    Domain classification is the task of mapping spoken language utterances to one of the natural language understanding domains in intelligent personal digital assistants (IPDAs). This is a major component in mainstream IPDAs in industry. Apart from official domains, thousands of third-party domains are also created by external developers to enhance the capability of IPDAs. As more domains are developed rapidly, the question of how to continuously accommodate the new domains still remains challenging. Moreover, existing continual learning approaches do not address the problem of incorporating personalized information dynamically for better domain classification. In this paper, we propose \newmodel, a neural network based approach for domain classification that supports incremental learning of new classes. Empirical evaluation shows that \newmodel\ achieves high accuracy and outperforms baselines by a large margin on both incrementally added new domains and existing domains.
\end{abstract}

\input{introduction}
\input{problem_definition}

\input{conda_solution}
\input{evaluation}
\input{related_work}
\input{conclusion}

\bibliography{naaclhlt2019}
\bibliographystyle{acl_natbib}

\end{document}

%% file: introduction.tex
\section{Introduction}
\label{sec:intro}
Domain classification is the task of mapping spoken language utterances to one of the natural language understanding (NLU) domains in intelligent personal digital assistants (IPDAs), such as Amazon Alexa, Google Assistant, and Microsoft Cortana, etc. \cite{sarikaya2017technology}. Here a domain is defined in terms of a specific application or functionality such as weather, calendar or music, which narrows down the scope of NLU. For example, given an utterance ``{\it Ask Uber to get me a ride}" from a user, the appropriate domain would be one that invokes the ``Uber" app.

Traditionally IPDAs have only supported dozens of well-separated domains, where each is defined in terms of a specific application or functionality such as calendar and weather \cite{sarikaya2016overview,tur2011spoken,el2014extending}. In order to increase the domain coverage and extend the capabilities of the IPDAs, mainstream IPDAs released tools to allow third-party developers to build new domains. Amazon’s Alexa Skills Kit, Google’s Actions and Microsoft’s Cortana Skills Kit are examples of such tools. To handle the influx of new domains, large-scale domain classification methods like \slmodel\ \cite{kim2018efficient} have been proposed and have achieved good performance.

As more new domains are developed rapidly, one of the major challenges in large-scale domain classification is how to quickly accommodate the new domains without losing the learned prediction power on the known ones. A straightforward solution is to simply retraining the whole model whenever new domains are available. However, this is not desirable since retraining is often time consuming. Another approach is to utilize continual learning where we dynamically evolve the model whenever a new domain is available. There is extensive work on the topic of continual learning, however there is very little on incrementally adding new domains to a domain classification system.

To mitigate this gap, in this paper we propose the \newmodel\ solution for continuous domain adaptation. Given a new domain, we keep all learned parameters, but only add and update new parameters for the new domain. This enables much faster model updates and faster deployment of new features to customers. To preserve the learned knowledge on existing domains to avoid the notorious catastrophic forgetting problem \cite{KemkerMAHK18}, we propose cosine normalization for output prediction and domain embedding regularization for regularizing the new domain embedding. Also, we summarize the data for existing domains by sampling exemplars, which will be used together with the new domain data for continuous domain adaptation. This is shown to further alleviate the overfitting on the new domain data. Empirical evaluation on real data with 900 domains for initial training and 100 for continuous adaptation shows that \newmodel\ out performs the baselines by a large margin, achieving 95.6\% prediction accuracy on average for the 100 new domains and 88.2\% accuracy for all seen domains after 100 new domains have been accommodated (only 3.6\% lower than the upperbound by retraining the model using all domain data). To summarize, we make the following contributions in this paper:
\begin{itemize}
    \item We introduce the problem of continuous domain adaptation for large-scale personalized domain classification.
    \item We describe \newmodel, a new solution for continuous domain adaptation with Cosine normalization, domain embedding regularization and negative exemplar sampling techniques. Our solution advances the research in continuous domain adaptation.
    \item We conduct extensive experiments showing that \newmodel\ achieves good accuracy on both new and existing domains, and outperforms the baselines by a large margin.
\end{itemize}

%% file: problem_definition.tex
\section{Background and Problem Definition}

\subsection{Domain Classification}
Domain classification is the task of mapping spoken language utterances to one of the NLU domains in IPDAs. A straightforward solution to tackle this problem is to ask users to explicitly mention the domain name with a specific invocation pattern. For example, for the utterance ``{\it Ask Uber to get me a ride}", the invocation pattern is ``{\it Ask \{domain\} to \{perform action\}}". While it makes things much simpler for the domain classifier, this significantly limits natural interaction with IPDAs as users need to remember the domain names as well as the invocation pattern. To address this limitation, {\it name-free domain classification} methods were developed for more user friendly interactions, and have been getting more attention recently. We specifically focus on the name-free scenario in this paper.

\subsection{The Shortlister System}
To our knowledge, the state-of-the-art for name-free domain classification is \slmodel\ \cite{kim2018efficient}, which leverages personalized user provided information for better classification performance. Specifically, it contains three main modules.

The first module is the LSTM-based encoder to map an utterance to a dimension-fixed vector representation. Given an utterance, each word is first represented as dense vectors using word embeddings, then a bidirectional LSTM \cite{graves2005framewise} is be used to encode the full utterance. 

The second module is the personalized domain summarization module. For each utterance from an IPDA user, a list of domains have been enabled by the user. These enabled domains can be viewed as user-specific personalized information. It has been shown that the domain classification accuracy can be significantly improved by leveraging information about enabled domains \cite{kim2018efficient}. To represent the domain enablement information, first each enabled domain is mapped to a fixed-dimensional embedding, then a summarization vector is generated by taking an attention weighted sum \cite{luong2015effective} over the enabled domain embeddings. 

Once the utterance representation and the enabled domain summarization are calculated, we concatenate the two vectors as the final representation. Then the third module, a feed-forward network, is used to predict the confidence score with a sigmoid function for each domain.



\subsection{Continuous Domain Adaptation}
As more new domains are developed, a major challenges in large-scale domain classification is quickly accommodating the new domains into the live production domain classification model without having to perform a full retrain. We refer to this problem as {\it Continuous Domain Adaptation} (CDA). In this paper, we specifically focus on the case of purely online learning where new domains where added one by one, since in practice we want to quickly integrate a new domain into the system as soon as it becomes available. We formally define the problem below.

\begin{definition}
\label{def:problem}
(Online continuous domain adaptation) Given a collection of $k$ domains $S_k=\{s_1, s_2, \dots, s_k\}$, suppose we have a dataset $\mathcal{D}_k$ defined on $S_k$ where each item is a triple $(u, s, E)$ with the utterance $u \in U$ (the set for all possible utterances), the ground-truth domain $s \in S_k$, and the enabled domains $E \subseteq S_k$. Denote $\mathcal{P}(S_k)$ as the powerset of $S_k$, a model $M_k: U \times \mathcal{P}(S_k) \to S_k$ has been trained on $\mathcal{D}_k$ for domain classification with the accuracy $M_k(\mathcal{D}_k)$. At some point, a new domain $s_{k+1}$ is available with the corresponding dataset $D_{k+1} = \{(u, s_{k+1}, E) \mid E \subseteq S_{k+1}\}$ with $S_{k+1} = S_{k}\cup \{s_{k+1}\}$. Taking advantage of $D_{k+1}$, the continuous adaptation for $s_{k+1}$ is to update $M_k$ to $M_{k+1}: U \times \mathcal{P}(S_{k+1}) \to S_{k+1}$ so that the model can make predictions for $s_{k+1}$, with the goal of maximizing $M_{k+1}(D_{k+1})$ and minimizing $M_{k}(\mathcal{D}_k) - M_{k+1}(\mathcal{D}_k)$.
\end{definition}

%% file: conda_solution.tex
\section{The CoNDA Solution}
We introduce \newmodel\ ({\bf Co}ntinuous {\bf N}eural {\bf D}omain {\bf A}daptation), a variation of \slmodel\ that is capable of handling online CDA decribed in Definition \ref{def:problem}. Similar to \slmodel, it has three main modules.

The first module is the LSTM-based utterance encoder which shares the same architecture as the one used in \slmodel, that maps an input utterance into a dense vector. After the training on the initial $k$-domain data $\mathcal{D}_k$, we freeze all parameters (i.e., the word embedding lookup and the bi-LSTM parameters) of this module from changing for the subsequent online domain adaptation tasks. Usually the value of $k$ is large enough (hundreds or even thousands in real-world, at least 100 in our experiments), thus it is safe to assume that the parameters have been tuned sufficiently well to encode utterances from all existing and future domains. In this work we treat new words in the new domains as unknown and leave the problem of vocabulary expansion as future work.

The second module is the personalized domain summarization module which will map the enabled domains of an input utterance to a dense vector representation. It is also similar to the one in \slmodel, except we will evolve the module as we are adding new domains. Specifically, given dataset $\mathcal{D}_k$ on $k$ domains for initial training, a domain embedding table $T_k \in \mathbb{R}^{k\times d_s}$ will be learned where $d_s$ is the size of the domain embeddings. When a new domain $s_{k+1}$ is available, we expand $T_{k}$ to $T_{k+1} \in \mathbb{R}^{(k+1)\times d_s}$ by: (1) freezing the learned embeddings for all known domains; (2) adding a new row $t_{k+1} \in \mathbb{R}^{d_s}$ to $T_k$ as the domain embedding for $s_{k+1}$ and updating the new parameters $t_{k+1}$ using all available training data at hand (i.e., the dataset $D_{k+1}$ and the negative samples which will be discussed later in this section). We repeat this procedure whenever a new domain is available. To avoid over-fitting on $t_{k+1}$, we introduce a new regularization term into the loss function. We describe the details in Section \ref{sec:domain_embedding_reg}.

The third module is a two-layer feed-forward network as the classifier. The first layer $f^{(1)}:\mathbb{R}^{d_u + d_s} \to \mathbb{R}^{d_h}$ maps the concatenation of the utterance embedding (in size $d_u$) and domain summarization (in size $d_s$) into fix-sized hidden representation (in size $d_h$) using a fully connected layer followed by SELU activation \cite{klambauer2017self}, which is identical to the one in \slmodel. Then the prediction layer $f^{(2)}:\mathbb{R}^{d_h} \to \mathbb{R}^{k}$ maps the hidden representation to the final domain prediction scores. Unlike \slmodel\ where the final prediction score is the dot product of the weight vector and the hidden representation, we choose to use the cosine score of the two, referred to as {\it cosine normalization}. To support online CDA when a new domain is available, we apply a similar approach to the domain embedding expansion described above to expand the prediction layer. Specifically, denote $W^{(2)}_k \in \mathbb{R}^{k\times d_h}$ be the weights for the prediction layer that has been trained on the initial $k$ domains. To adapt the new domain $d_{k+1}$, we expand $W^{(2)}_k$ to $W^{(2)}_{k+1} \in \mathbb{R}^{(k+1)\times d_h}$ by first freezing all learned parameters and adding a new row of learnable parameters $w_{k+1} \in \mathbb{R}^{d_h}$ to $W^{(2)}_k$.

As each time we only add one new domain, all training utterances during the update will have the same label. Thus, it's easy to overfit the new data such that catastrophic forgetting occurs. Inspired by \cite{rebuffi2017icarl}, we also propose a negative sampling procedure to leverage (limited) information on the known domains to alleviate the catastrophic forgetting problem. For the rest of the section, we will first talk about cosine normalization, and then domain embedding regularization, and finally negative sampling.

\subsection{Cosine Normalization}
\begin{figure}[t]
\centering
\includegraphics[width=\columnwidth]{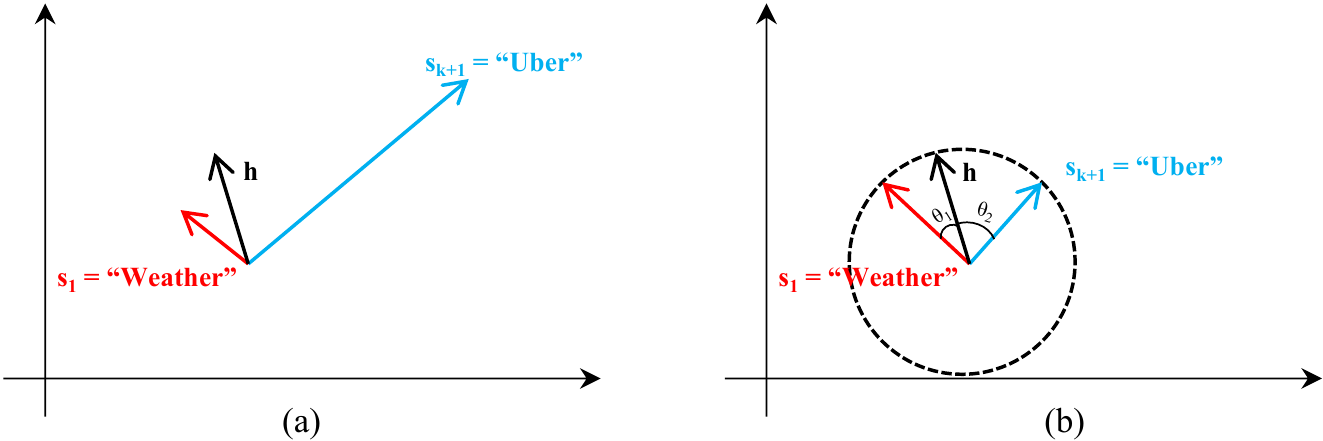}
\caption{Cosine Example.}\label{fig:cosine_example}
\end{figure}

As mentioned above, we use the cosine similarity of the weights and the hidden representation vector instead of the linear dot product in the prediction layer. Formally, let $f^{(2)}_k: \mathbb{R}^{d_h} \to [-1, 1]^{k}$ be the prediction layer for $k$ domains with parameters $W^{(2)}_k \in \mathbb{R}^{k\times d_h}$. Given an input hidden representation $h \in \mathbb{R}^{d_h}$ from $f^{(1)}$, the score for the $i$-th domain under cosine normalization is:

\begin{equation}
\label{eq:cos}
    f^{(2)}_{k, i}(h) = \cos{\left(h, W^{(2)}_{k,i}\right)} = \frac{h \cdot W^{(2)}_{k,i}}{\norm{h}\norm{W^{(2)}_{k,i}}}
\end{equation}

To understand why cosine is better in the case of online CDA, let's first see the problem with the dot-product method. Suppose we are accommodating $s_{k+1}$ with dataset $D_{k+1}$, because we train the new parameters $w_{k+1}$ only on $D_{k+1}$ where all utterances have the same domain $s_{k+1}$, the model can easily get good training performance on $M_{k+1}(D_{k+1})$ by simply maximizing the values in $w_{k+1}$ such that the dot product of the hidden representation with $w_{k+1}$ is larger than the dot product with any other $w_{i}, 1 \leq i \leq k$. Effectively this leads to the model predicting domain $s_{k+1}$ for any given utterance. Using cosine normalization instead as described in Eq. \ref{eq:cos} removes the incentive to maximize the vector length of $w_{k+1}$.

\begin{example}
\label{ex:cosine_norm}
Suppose $M_k$ has been initially trained on $\mathcal{D}_k$, and domain $s_1$=``Weather". Given an utterance $u$ = ``What's the weather today?", $M_k$ correctly classifies $u$ into $s_1$. Now a new domain $s_{k+1}$=``Uber" is coming and we evolve $M_{k}$ to $M_{k+1}$. As the norm of the weights $w_{k+1}$ could be much larger than $w_1$ in the prediction layer, even if the hidden representation $h$ of $u$ is closer to $s_1$ in direction, $M_{k+1}$ will classifier $u$ into $s_{k+1}$ as it has a higher score, shown in Figure \ref{fig:cosine_example}.a. However if we measure the cosine similarity, $M_{k+1}$ will classify $u$ correctly because we now care more about the directions of the vectors, and the angle $\theta_1$ between $h$ and $s_1$ is smaller (representing higher similarity) than the angle $\theta_2$ between $h$ and $s_{k+1}$, as shown in Figure \ref{fig:cosine_example}.b.
\end{example}

As we use the cosine normalization, all prediction scores are mapped into the range [-1, 1]. Therefore it's not proper to use log-Sigmoid loss function as in \slmodel. So accompanying with the cosine normalization, the following hinge loss function has been used instead:
\begin{dmath}
\label{eq:cos_hinge}
    \mathcal{L}_{hinge}=\sum_{i=1}^ny_i\max\{\Delta_{pos}-o_i, 0\} + \sum_{i=1}^n(1-y_i)\max\{o_i-\Delta_{neg}, 0\}
\end{dmath}
where $n$ is the number of all domains, $o_i$ is the predicted score for each domain, $y$ is a $n$-dimensional one-hot vector with 1 in the ground-truth label and 0 otherwise. $\Delta_{pos}$ and $\Delta_{neg}$ are the hinge thresholds for the true and false label predictions respectively. The reason we use hinge loss here is that it can be viewed as another way to alleviate the overfitting on new data, as the restrictions are less by only requiring the prediction for the ground-truth to be above $\Delta_{pos}$ and false domain predictions below $\Delta_{neg}$. Our experiments show that this helps the model get better performance on the seen domains.

\subsection{Domain Embedding Regularization}
\label{sec:domain_embedding_reg}
In this section, we introduce the regularizations on the domain embeddings used in the personalized domain summarization module. Recall that given an utterance $u$ with $h^u$ as the hidden representation from the encoder and its enabled domains $E$, personalized domain summarization module first compares $u$ with each $s_i \in E$ (by calculating the dot product of $h^u$ and the domain embedding $t_i$ of $s_i$) to get a score $a_i$, then gets the weight $c_i = \exp{(a_i)}/\sum_{a_j}\exp{(a_j)}$ for domain $s_i$, and finally computes the personalized domain summary as $\sum_{e_i \in E}c_i \cdot t_i$. We observed that after training on the initial dataset $D_k$, the domain embedding vectors tend to roughly cluster around a certain (random) direction in the vector space. Thus, when we add a new domain embedding $s_{k+1}$ to this personalization module, the model tends to learn to move this vector to a different part of the vector space such that its easier to distinguish the new domain from all other domains. Moreover, it also increases the $\ell_2$ norm of the new domain embedding $t_{k+1}$ to win over all other domains.

\begin{example}
\label{ex:der}
Suppose a similar scenario to Example \ref{ex:cosine_norm} where we have $s_1$ = ``Weather" in $S_{k}$ and a new domain $s_{k+1}$ = ``Uber". As most utterances in $D_{k+1}$ have $s_{k+1}$ as an enabled domain, it's easy for the model to learn to enlarge the norm of the new domain embedding $t_{k+1}$ as well as make it close to the context of ride sharing, so that $t_{k+1}$ can dominate the domain summarization. Then coordinating with the new weights $w_{k+1}$ in the prediction layer $f^{(2)}_{k+1}$, the network can easily predict high scores $s_{k+1}$ and fit the dataset $D_{k+1}$. However, when we have utterances belonging to $s_1$ with $s_{k+1}$ as an enabled domain, $s_{k+1}$ may still dominate the summarization which makes the prediction layer tends to cast those utterances to $s_{k+1}$. We don't observe this on the initial training on $\mathcal{D}_k$ because $s_{k+1}$ was not visible at that time, thus cannot be used as an enabled domain. And it's even worse if $s_1$ is similar to $s_{k+1}$ in concept. For example if $s_1$ = ``Lyft", in this case the utterances of the two domains are also similar, making the dot product of $t_{k+1}$ and the hidden representations of the $s_1$'s utterances even larger.
\end{example}

To alleviate this problem, we add a new domain embedding regularization term in the loss function to constrain the new domain embedding vector length and force it to direct to a similar area where the known domains are heading towards, so that the new domain will not dominate the domain summarization. Specifically,
\begin{dmath}
\label{eq:der_loss}
    \mathcal{L}_{der}=\sum_{i=1}^k\lambda_i\max\{\Delta_{der} - \cos(t_{k+1}, t_{i}), 0\} + \frac{\lambda_{norm}}{2}\norm{t_{k+1}}^2
\end{dmath}

We call the first part of Eq. \ref{eq:der_loss} on the right hand side as the {\it domain similarity loss} where we ask the new domain embedding $t_{k+1}$ to be similar to known domain $t_i$'s controlled by a Cosine-based hinge loss. As we may not need $t_{k+1}$ to be similar to all seen domains, a coefficient $\lambda_i$ is used to weight the importance each similarity loss term. In this paper we encourage $t_{k+1}$ to be more similar to the ones sharing similar concepts (e.g. ``Uber" and ``Lyft"). We assume all training data are available to us, and measure the similarity of two domains by comparing their average of utterance hidden representations. 

Specifically, denote $\varphi: U \to \mathbb{R}^{d_u}$ as the LSTM-encoder that will map an utterance to its hidden representation with dimension $d_u$. For each domain $s_i \in S_{k+1}$, we first calculate the average utterance representation on $D_i$
\begin{equation}
\label{eq:utt_ave}
    \widetilde h_{i} = \sum_{(u, s_i, e)\in D_{i}}\frac{\varphi(u)}{|D_i|}
\end{equation}
Then we set $\lambda_i = \lambda_{dsl} \max\{\cos(\widetilde h_i, \widetilde h_{k+1}), 0\}$ with $\lambda_{dsl}$ as a scaling factor.

Combining Eq. \ref{eq:cos_hinge} and \ref{eq:der_loss}, the final loss function for optimization is: 
$
    \mathcal{L}_{total} = \mathcal{L}_{hinge} + \mathcal{L}_{der}
$

\begin{figure*}[t]
\centering
\includegraphics[width=\textwidth]{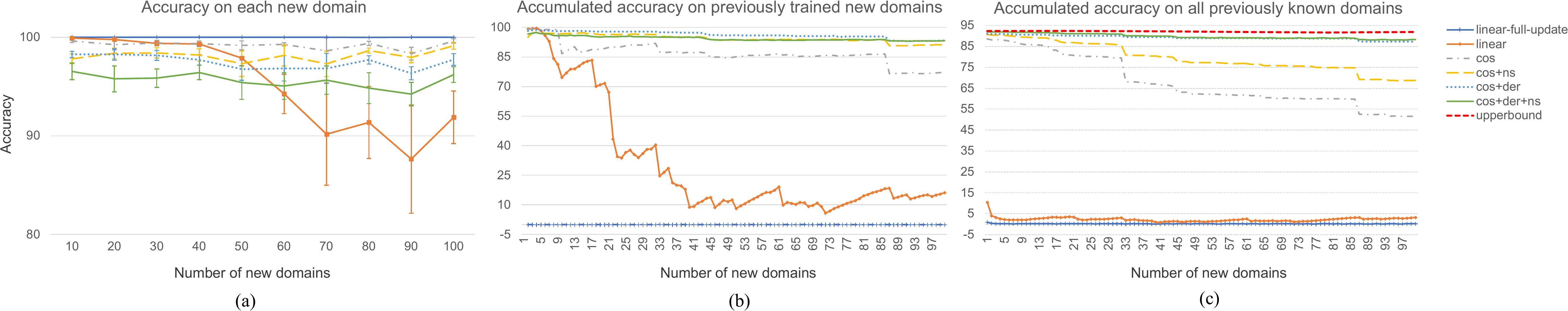}
\vspace{-6mm}
\caption{Overall evaluation. (a) shows the accuracy for new domains. (b) shows the accumulated accuracy for previous new domains that have been adapted to the model so far. (c) shows the accumulated accuracy for all known domains including the ones used for initial training and all previously adapted new domains.}\label{fig:overall_perf}
\end{figure*}

\begin{figure*}[t]
\centering
\includegraphics[width=\textwidth]{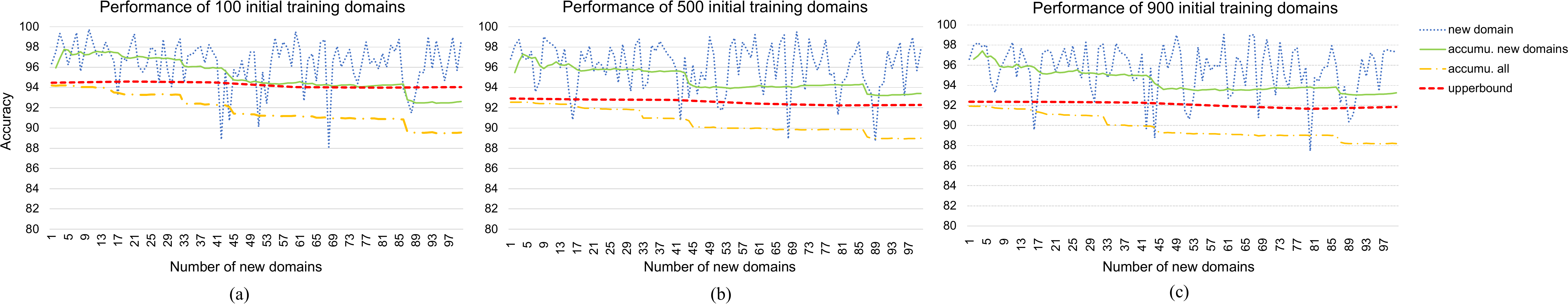}
\vspace{-3mm}
\caption{The model performance on different number of initial training domains. The red dashed line shows the upperbound of the accumulated accuracy, which is generated by retraining the model on all domains seen so far.}\label{fig:diff_num_init_domains}
\vspace{-1mm}
\end{figure*}

\subsection{Sampling Negative Exemplars}
So far we developed our method by training only on the new data $D_{k+1}$, and use regularizations to prevent overfitting. However, in many real applications all of the training data, not only $D_{k+1}$, is actually available, but it's not affordable to retrain the full model using all data. Inspired by \cite{rebuffi2017icarl}, we can select a set of exemplars from the previously trained data to further improve continual adaptation.

Suppose we are handling the new domain $s_{k+1}$ with $D_{k+1}$, and all data trained previously is $\mathcal{D}_k$ on $k$ domains $S_{k}$. For each known $s_i \in S_{k}$, we pick $N$ utterances from $D_i$ as the exemplars for $s_i$. Denote $P_i$ be the exemplar set for $s_i$ and $P=\bigcup_{i=1}^kP_i$ be the total exemplar set. To generate each $P_i$, we pick the top-$N$ utterances that are closest to the average of the utterance hidden representation. Specifically, following Eq. \ref{eq:utt_ave}, we first get the average representation $\widetilde h_i$, then $P_i$ is defined as follow:
\begin{equation}
\label{eq:neg_sampling}
    P_i=\argmax_{P_i\subseteq D_i, |P_i|=N}\sum_{(u, s_i, e) \in P_i}\cos\left(\varphi(u), \widetilde h_i\right)
\end{equation}
If multiple candidates satisfying Eq. \ref{eq:neg_sampling} for $P_i$, we randomly pick one as $P_i$ to break the tie. Once the domain adaptation for $s_{k+1}$ is done, we similarly generate $P_{k+1}$ and merge it to $P$. We repeat this procedure for negative sampling whenever a new domain is coming later. 

As we add more new domains, the exemplar set $P$ also grows. For some new domain $D_{k+1}$, we may have $|P| \gg |D_{k+1}|$. In this case, the prediction accuracy on the new domain data could be very low as the model will tend to not making mistakes on $P$ rather than fitting $D_{k+1}$. To alleviate this problem, when $|P| > |D_{k+1}|$, we select a subset $P' \subseteq P$ with $|P'| = |D_{k+1}|$, and $P'$ will be used as the final exemplar set to train together with $D_{k+1}$. To generate $P'$, we just randomly sample a subset from $P$, since it was observed to be effective in our experiments.


%% file: evaluation.tex
\section{Empirical Evaluation}
\subsection{Experiment Setup}
\paragraph{Dataset:} We use a dataset defined on 1000 domains for our experiments which has 2.53M utterances, and we split them into two parts. The first part contains 900 domains where we use it for the initial training of the model. It has 2.06M utterances, and we split into training, development and test sets with ratio of 8:1:1. We refer to this dataset as ``InitTrain". The second part consists of 100 domains and is used for the online domain adaptation. It has 478K utterances and we split into training, development and test sets with the same 8:1:1 ratio. We refer to this dataset as ``IncTrain".

\paragraph{Training Setup:} We implement the model in PyTorch \cite{paszke2017automatic}. All of the experiments are conducted on an Amazon AWS p3.16xlarge\footnote{https://aws.amazon.com/ec2/instance-types/p3/} cluster with 8 Tesla V100 GPUs. For initial training, we train the model for 20 epochs with learning rate 0.001, batch size 512. For the continuous domain adaptation, we add the new domains in a random order. Each domain data will be trained independently one-by-one for 10 epochs, with learning rate 0.01 and batch size 128. For both training procedures, we use Adam as the optimizer. The development data is used to pick the best model in different epoch runs. We evaluate the classification accuracy on the test set.

\subsection{Overall Performance}
We first talk about the overall performance. In our experiments we select two baselines. The first one \texttt{linear-full-update} which simply extends \slmodel\ by adding new parameters for new domains and conducting full model updating. The second \texttt{linear} is similar to the first baseline except that we freeze all trained parameters and only allow new parameter updating. Both the two baselines update the model with $D_{k+1}$ dataset only. To show the effectiveness of each component of \newmodel, we choose four variations. The first one is \texttt{cos} where we apply the Cosine Normalization (CosNorm). The second one \texttt{cos+der} applies CosNorm with the domain embedding regularization. The third one \texttt{cos+ns} uses both CosNorm and negative exemplars. And the last one \texttt{cos+der+ns} is the combination of all three techniques, which is our \newmodel\ model. For hyperparameters, we pick $\Delta_{pos}=0.5, \Delta_{neg}=0.3, \Delta_{der}=0.1, \lambda_{dsl}=5$, and $\lambda_{norm}=0.4$.

Figure \ref{fig:overall_perf} shows the accuracy for new domain adaptations. From the figure, here are the main observations. First, without any constraints, \texttt{linear-full-update} can easily overfits the new data to achieve 100\% accuracy as shown in Figure \ref{fig:overall_perf}(a), but it causes catastrophic forgetting such that the accuracy on seen domains is (almost) 0 as shown in Figure \ref{fig:overall_perf}(b) and (c). By freezing the all trained parameters, the catastrophic forgetting problem is a bit alleviated for \texttt{linear}, but the accuracy on the seen domains is still very low as we add more new domains. Second, \texttt{cos} produces much better accuracy on seen domains with a bit lower accuracy on each new domain, showing the effectiveness of the Cosine normalization. Third, as we add more regularizations to the model, we get better accuracy on the seen domains (Figure \ref{fig:overall_perf} (b) and (c)), at the cost of sacrificing a bit on the new domain accuracy (Figure \ref{fig:overall_perf} (a)). Also, \texttt{cos+der+ns} (the \newmodel\ model) achieves the best performance, with an average of 95.6\% accuracy for each new domain and 88.2\% accuracy for all previously seen domains after we add 100 new ones, which is only 3.6\% lower than the upperbound (by retraining the model on the whole dataset). These demonstrate the superiority of our method.

\subsection{Micro-benchmarks}

\paragraph{Using Different Number of Initial Domains:} We vary the number of domains for initial training to see if it will have a big impact on the model performance. Specifically, we pick 100 and 500 domains from InitTrain, and use the same IncTrain data for domain adaptation. Figure \ref{fig:diff_num_init_domains} compares the model performance on these three different number (i.e., 100, 500, 900) of initial training domains. From the figure we can see that the curves share a similar pattern regardless of the number of initial domains, showing that our model is stable to the number of domains used for initial training.

\begin{figure*}[t]
\centering
\includegraphics[width=0.9\textwidth]{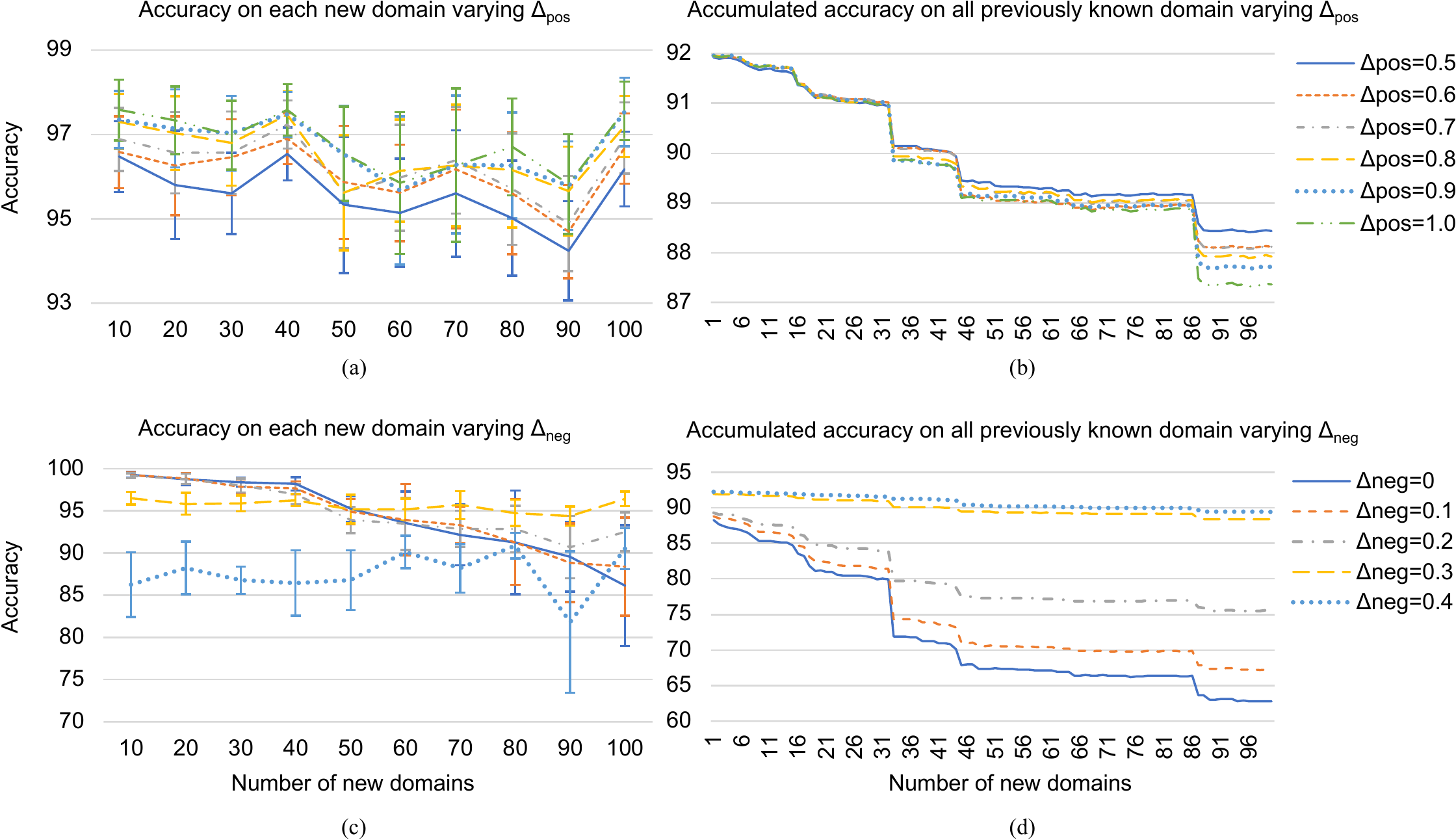}
\caption{Model performance by varying the hinge loss thresholds. (a) and (b) shows accuracy on new domain and accumulated accuracy for all seen domains respectively by varying $\Delta_{pos}$. Similarly, (c) and (d) shows the accuracy performance by varying $\Delta_{neg}$.}\label{fig:hinge_loss_thres}
\end{figure*}

\paragraph{Varying the hinge loss thresholds:} We vary the classification hinge loss thresholds $\Delta_{pos}$ and $\Delta_{neg}$ to see how it will affect the performance. Specifically, we fix $\Delta_{neg}=0.3$ and vary $\Delta_{pos}$ from 0.5 to 1.0, and fix $\Delta_{pos}=0.5$ and vary $\Delta_{neg}$ from 0 to 0.4, respectively. For both of the them we use 0.1 as the step size. Figure \ref{fig:hinge_loss_thres} shows the model performance. From the figures, we summarize the following observations. First, as we increase $\Delta_{pos}$, on average the accuracy on each new domain gets better (Figure \ref{fig:hinge_loss_thres}(a)), but we loss performance on all seen domains (Figure \ref{fig:hinge_loss_thres}(b)). This is in accord with our intuition that a larger $\Delta_{pos}$ puts more constraint on the new domain predictions such that it tends to overfit the new data and exacerbates catastrophic forgetting on existing domains. Second, as we increase $\Delta_{neg}$, on average the accuracy on each new domain gets worse (Figure \ref{fig:hinge_loss_thres}(c)), but we get better performance on existing domains. This is because a larger $\Delta_{neg}$ narrows down the prediction ``margin" between positive and negative domains (similar to decreasing $\Delta_{pos}$), so that less constraint has been put onto predictions to alleviate overfitting on the new domain data.

\begin{figure}[t]
\centering
\includegraphics[width=0.9\columnwidth]{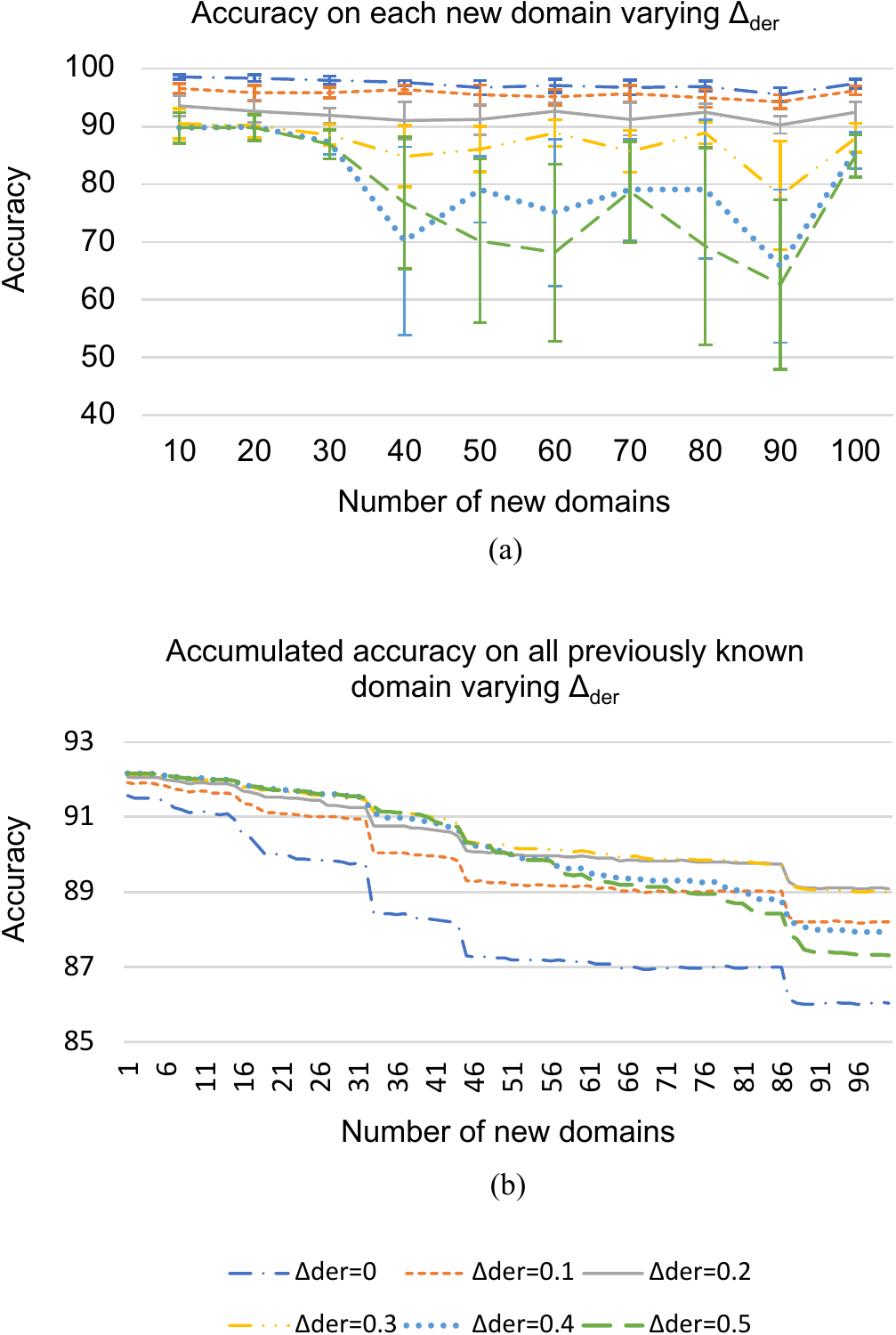}
\caption{Model performance by varying $\Delta_{der}$. (a) shows the accuracy on new domains, and (b) shows the accumulated accuracy for all seen domains.}\label{fig:der_thres}
\end{figure}

\paragraph{Varying the domain similarity loss threshold:} We vary the threshold $\Delta_{der}$ to see how it will affect the model performance. Specifically, we vary $\Delta_{der}$ from 0 to 0.5 with step size 0.1, and Figure \ref{fig:der_thres} shows the model performance. As we increase $\Delta_{der}$, the performance on the new domains gets worse, and the drop is significant when $\Delta_{der}$ is large. On the other hand, the accumulated accuracy on seen domains increases when we start to increase $\Delta_{der}$, and drops when $\Delta_{der}$ is too large. This means we when we start to make the new domain embeddings to be similar to the existing ones, we alleviate the problem that the new domain dominates the domain summarization. Thus the accuracy on existing domains improves at the cost of sacrificing some accuracy on the new domains. However, if we continue to increase $\Delta_{der}$ to make it very similar to some of existing domains, the new domain will compete with some existing ones so that we loss accuracy on both new and existing domains.

\paragraph{Varying the weights for domain similarity loss:} To see how the weighted domain similarity loss will affect the performance, we compare it against the plain version without the utterance similarity weights. Specifically, we set each $\lambda_i = \lambda_{dsl}$ having the same value. And our experiments show that the plain version gets the average accuracy 94.1\% on the new domains, which is 1.5\% lower than the weighted version, and 88.7\% accumulated accuracy on all domains after adding 100 new domains, which is 0.5\% higher than the weighted version. This means we can get a bit higher accumulated accuracy at the cost of sacrificing more new domain accuracy. In real applications, the decision to whether use weighted domain similarity loss should be made by trading off the importance of the new and existing domains.

\paragraph{Varying the number of used negative exemplars:} As we mentioned before, we down-sample the negative exemplar set $P$ to reduce the impact on new domain performance. To see if it's necessary, we compare it against the one without down-sampling. Our experiments show that without down-sampling, the model achieves 87.5\% new domain accuracy on average which is 8.1\% lower than the down-sampling version, and 87.2\% accumulated accuracy on all domains which is 1.0\% lower than the down-sampling one. This means down-sampling effectively improve the model performance.

\begin{table}[t]
\centering
\includegraphics[width=\columnwidth]{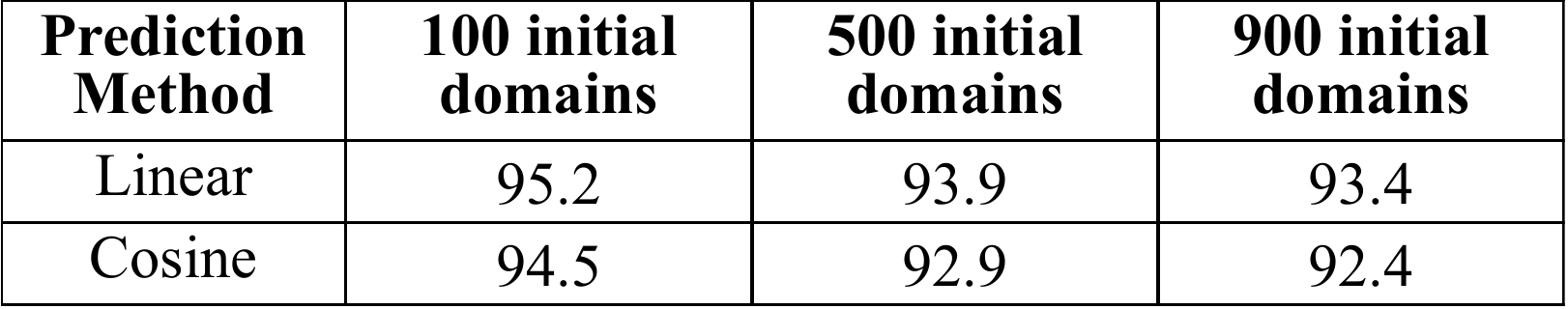}
\caption{Linear dot product versus Cosine normalization on initial training for different number of domains.}\label{table:cos_vs_linear}
\end{table}

\paragraph{Effect of Cosine normalization in initial training:} We have shown Cosine normalization with hinge loss works better than linear dot product with sigmoid loss (used in \slmodel) for CDA. Here we compare the two on the regular training setting where we train the model from scratch on a large dataset. Specifically, we compare the initial training performance on 100, 500, and 900 domains which are the same as we used earlier. Table \ref{table:cos_vs_linear} shows the accuracy numbers. From the table we see that Linear works better than Cosine by 0.7-1.0\% across different number of domains. Though the difference is not large, this means Linear could be a better option than Cosine when we train the model from scratch.

\paragraph{Varying the order of the new domains:} To see if the incoming order of the new domains will affect the performance, we generate two different orders apart from the one used in overall evaluation. The first one sorts the new domains on the number of utterances in the decreasing order, and the second in the increasing order. Denote these three orders as ``random", ``decreasing", and ``increasing", and we conduct domain adaptation on these orders. Our experiments show that they achieve 95.6\%, 95.5\%, and 95.6\% average accuracy on new domains respectively, and 88.2\%, 88.2\%, and 88.1\% accumulated accuracy on all domains after accommodating all 100 new domains. This indicates that there is no obvious difference on model performance, and our model is insensitive to the order of the new domains.

\begin{figure}[t]
\centering
\includegraphics[width=\columnwidth]{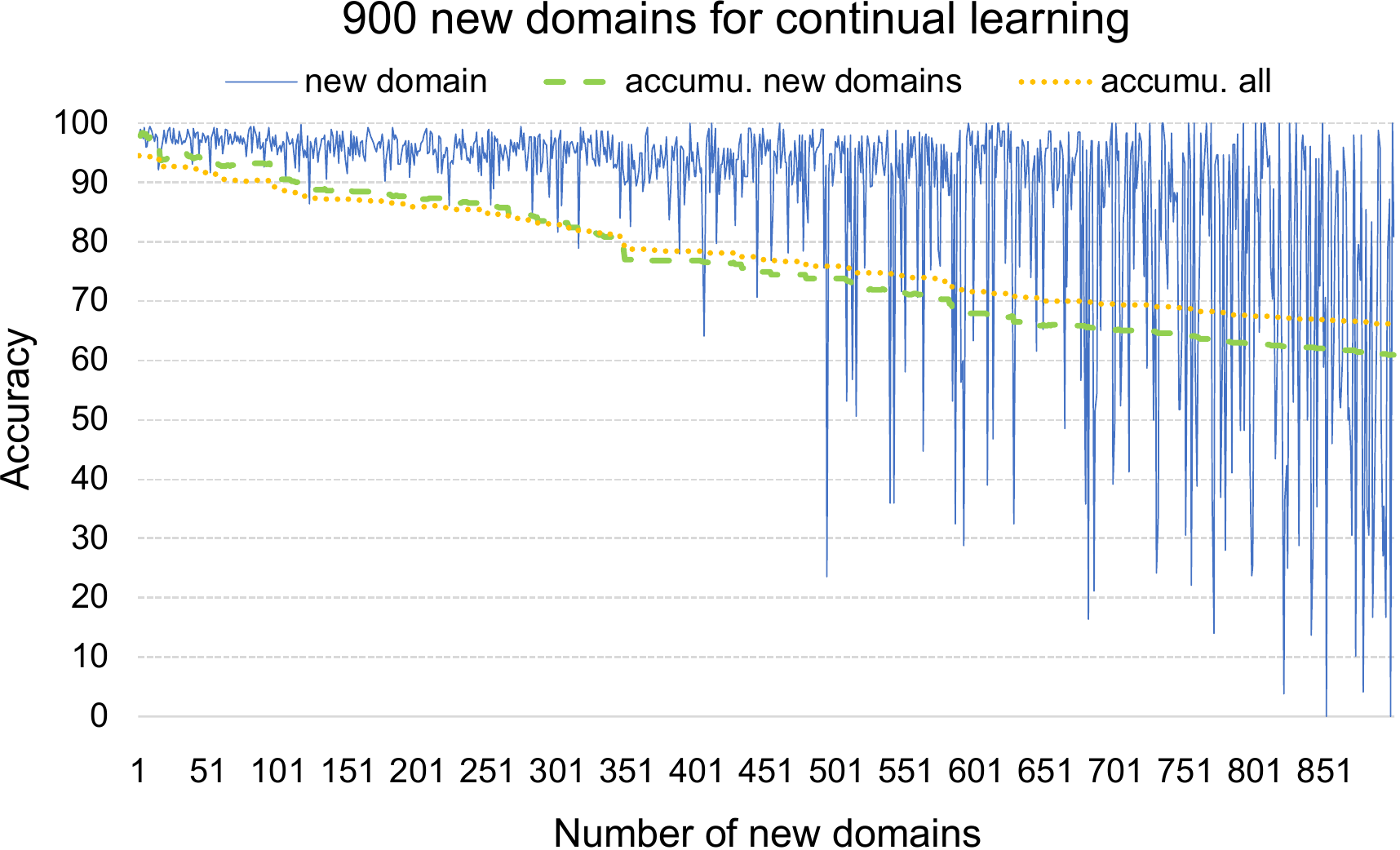}
\caption{Using 900 new domains for continual learning.}\label{figure:more_new_domains}
\vspace{-10pt}
\end{figure}

\paragraph{Using more new domains:} We also experimented with adding a large number of new domains to see the limit of \newmodel. Figure \ref{figure:more_new_domains} shows the results by continuously adapting 900 new domains one-by-one. From the figure we can see that at the early stage of the new domain adaptation (e.g., first 200 new domains), we get high new domain accuracy with little performance decrease on the existing domains. After that, the new domain performance becomes more unstable with violent oscillation, and the existing domain accuracy decreases more quickly. This suggests that we cannot run the new domain adaptation forever, and after adapting a certain number of new domains (e.g., 200 new domains), it's more preferable to train the whole model from scratch.

%% file: related_work.tex
\section{Related Work}
\paragraph{Domain Classification:} Traditional domain classifiers were built on simple linear models such as Multinomial logistic regression or Support Vector Machines \cite{tur2011spoken}. They were typically limited to a small number of domains which were designed by specialists to be well-separated. To support large-scale domain classification, \cite{kim2018efficient} proposed \slmodel, a neural-based model. \cite{KimKKS18} extended \slmodel\ by using additional contextual information to rerank the predictions of \slmodel. However, none of them can continuously accommodate new domains without full model retrains.

\paragraph{Continuous Domain Adaptation:} To our knowledge, there is little work on the topic of continuous domain adaptation for NLU and IPDAs. \cite{KimSK17} proposed an attention-based method for continuous domain adaptation, but it introduced a separate model for each domain and therefore is difficult to scale.

\paragraph{Continual Learning:} Several techniques have been proposed to mitigate the catastrophic forgetting \cite{KemkerMAHK18}. Regularization methods add constraints to the network to prevent important parameters from changing too much\ \cite{kirkpatrick2017overcoming,ZenkePG17}. Ensemble methods alleviate catastrophic forgetting by explicitly or implicitly learning multiple classifiers and using them to make the final predictions \cite{dai2009eigentransfer,ren2017life,fernando2017pathnet}. Rehearsal methods use data from existing domains together with the new domain data being accommodated to mitigate the catastrophic forgetting \cite{robins1995catastrophic,draelos2017neurogenesis,rebuffi2017icarl}. Dual-memory methods introduce new memory for handling the new domain data \cite{gepperth2016bio}. Among the existing techniques, our model is most related to the regularization methods. However, unlike existing work where the main goal is to regularize the learned parameters, we focus on regularizations on the newly added parameters. Our model also shares similar ideas to \cite{rebuffi2017icarl} on the topic of negative exemplar sampling. 

%% file: conclusion.tex
\section{Conclusion and Future Work}
In this paper, we propose \newmodel\ for continuous domain adaptation. By using various normalization and regularizations, our model achieves high accuracy on both the accommodated new domains and the existing known domains, and outperforms the baselines by a large margin. For future work, we consider extending the model to handle unknown words. Also, we want to find a more principled way to down sample the negative exemplars.